\documentclass{article}

\usepackage{comment}
\usepackage{enumitem}

\setlist[itemize]{leftmargin=4mm}

\usepackage[preprint]{neurips_2019}

\usepackage[utf8]{inputenc} 
\usepackage[T1]{fontenc}    
\usepackage{hyperref}       
\usepackage{url}            
\usepackage{booktabs}       
\usepackage{amsfonts}       
\usepackage{nicefrac}       
\usepackage{microtype}      
\usepackage{amsmath, mathtools, amsthm, amssymb}
\usepackage{float}
\usepackage{pdfpages}

\usepackage[compact]{titlesec}         

\usepackage{tabularx}

\usepackage{algorithm}
\usepackage{algorithmic}

\usepackage{graphicx}
\usepackage{subfig}
\usepackage{authblk}

\DeclareMathOperator{\EX}{\mathbb{E}}

\title{Avoidance Learning Using Observational Reinforcement Learning}

\author{%
  {\textbf{David Venuto\textsuperscript{1,2}, Léonard Boussioux\textsuperscript{2,3,4,5}, Junhao Wang\textsuperscript{1,2}, Rola Dali\textsuperscript{1},} \\ \textbf{Jhelum Chakravorty\textsuperscript{1,2}, Yoshua Bengio\textsuperscript{2,4,6}, Doina Precup\textsuperscript{1,2,7}}} \\
  \textsuperscript{1}McGill University, \textsuperscript{2}Mila, 
  \textsuperscript{3}MIT, \textsuperscript{4}Université de Montréal,
  \textsuperscript{5}École CentraleSupélec,\\
  \textsuperscript{6}CIFAR Senior Fellow, \textsuperscript{7}DeepMind\\
  \texttt{\{david.venuto,junhao.wang,rola.dali,jhelum.chakravorty\}@mail.mcgill.ca,  leobix@mit.edu, yoshua.bengio@mila.quebec, dprecup@cs.mcgill.ca} \\
}

\begin{document}

\maketitle

\begin{abstract}
  Imitation learning seeks to learn an expert policy from sampled demonstrations. However, in the real world, it is often difficult to find a perfect expert and avoiding dangerous behaviors becomes relevant for safety reasons. We present the idea of \textit{learning to avoid}, an objective opposite to imitation learning in some sense, where an agent learns to avoid a demonstrator policy given an environment. We define avoidance learning as the process of optimizing the agent's reward while avoiding dangerous behaviors given by a demonstrator. In this work we develop a framework of avoidance learning by defining a suitable objective function for these problems which involves the \emph{distance} of state occupancy distributions of the expert and demonstrator policies. We use density estimates for state occupancy measures and use the aforementioned distance as the reward bonus for avoiding the demonstrator. We validate our theory with experiments using a wide range of partially observable environments. Experimental results show that we are able to improve sample efficiency during training compared to state of the art policy optimization and safety methods.
\end{abstract}

\section{Introduction}
It is in the nature of living organisms to harness the knowledge of others who are more experienced than them to develop behaviors and skills that are crucial for tasks throughout their life \citep{nehaniv2007imitation} and humans are no exception. Much of this skill acquisition is done in an observational process in which we observe the behaviors of other agents and imitate them. In scenarios that arise in Safe Reinforcement Learning (Safe RL), one wishes to avoid exploring \emph{risky} behaviors while pursuing a goal. For example, it is desirable for an autonomous vehicle to avoid digressing to sideways or colliding with other vehicles or pedestrians.  We wish to address \emph{Avoidance Learning} -- a problem reported and extensively studied in human behavior~\citep{Turnwald2016, cite2} -- and mathematically formulate the corresponding learning problem for an artificial agent.

Learning from an expert is a well-studied concept in RL and robotics \citep{argall2009survey}.  It can be categorized into two main approaches: Behavior Cloning \citep{Sammut2010} and Inverse Reinforcement Learning \citep{abbeel2004apprenticeship}. In the former case the agent tries to mimic the policy of an expert in a supervised fashion, whereas in the latter case, it recovers a reward function from the expert to optimize its policy. A recent inverse reinforcement learning algorithm is Generative Adversarial Imitation Learning (GAIL) \citep{ho2016generative}, where the reward of the expert is estimated by the agent in a two-player zero-sum game setting: a generator tries to approximate the expert policy while a discriminator distinguishes between expert and novice behaviors.

A natural question that arises in this premise of \emph{learning from observation} is how to leverage the information obtained from observing a \emph{dangerous} demonstrator to maximize the agent's goal (Observational Learning) \citep{Borsa2017ObservationalLB}. 
One can pose such a situation as an \emph{anti-imitation learning} problem, where the underlying constrained optimization problem is to maximize the agent's reward while staying \emph{as far as possible} from the demonstrator's observations. 

In the real world, circumstances may arise where an agent trying to learn from a demonstrator lacks the precise knowledge of the demonstrator's actions and can only observe his states (i.e., the consequence of his actions). In addition, the objective of the demonstrator is often  unknown to the agent (as in the case of inverse RL). All the agent can observe is the changes of \textit{state} distribution of the demonstrator throughout the learning process.  Many safe RL methods require either an explicit constraint on the policy \citep{pmlr-v70-achiam17a} or manage risk by reducing measures of variability in cost such as the conditional value-at-risk (CVaR) \citep{Tamar2014PolicyGB}. 


In this work, we present an \emph{avoidance learning} (AvL) method that requires only knowledge of the \emph{state-only} trajectories from the demonstrator and no explicit measure of danger or policy constraints, only demonstrations. The learning problem is a constrained optimization problem where the agent maximizes the discounted sum of rewards while trying to maximize the divergence of its own estimated state occupancy distribution from that of the demonstrator.  This allows the agent to optimize its own policy while avoiding a bad demonstrator's policy. Unlike many state-of-the-art methods in safe RL, our method does not require explicit engineering of negative rewards or constraints.

Estimating the state occupancy distribution of the demonstrator's policy is not trivial when the size of the state space becomes large and a simple count-based estimate does not suffice. Thus, we first present a formulation of this problem using a Variational Auto-Encoder (VAE) \citep{kingma2013auto} to estimate the stationary distributions of the agent's and demonstrator's state occupancy distributions.  We propose a novel objective function for the constrained optimization problem inherent to AvL, which is the sum of the agent's reward and the  Kullback-Leibler (KL) divergence between these two stationary distributions. This is essentially the \emph{Lagrangian relaxation} (also known as the Lagrangian dual) problem of the corresponding the original constrained optimization problem and the agent's optimal policy is the solution of maximizing the Lagrangian relaxation objective.

We apply our proposed method for AvL to both 2D and 3D partially observable environments. We use a convolutional neural density model trained on demonstrator state sequences to find an \textit{avoidance} reward bonus for agent training. Our experimental results corroborate our theory that this method successfully learns a policy that avoids the dangerous demonstrator trajectories while still finding the optimal reward. Furthermore, this method results in faster learning for the agent trained with the novel objective by avoiding exploration of \emph{dangerous} states.
\section{Preliminaries}
A Markov Decision Process (MDP) is made of a tuple $\langle S, \mathcal{A}, \mathbb P, R, \gamma \rangle$ where $S$ is a set of states, $\mathcal{A}$ is the set of actions available to the agent, $\mathbb P$ is the transition kernel giving a probability over next states given the current state and action, $R : S \times \mathcal{A} \rightarrow R$ is a reward function and $\gamma \in (0, 1)$ is a discount factor. $s_t$ and $a_t$ are respectively the state and action of the expert at time instant $t$. We define a policy $\pi$ as the probability distribution over actions conditioned on the current state; $\pi : S \times \mathcal{A} \rightarrow [0,1]$. The value of a policy is defined as $V_\pi(s) = \EX_\pi [\sum_{t=0}^{\infty} \gamma^t r_{t+1} | s]$, where $\mathbb E$ denotes the expectation. The entropy of a policy is $H(\pi) = - \EX[\log( \pi(a_t|s_t))]$.
An agent follows a policy $\pi$ and receives reward from the environment. A state-action value function is $Q_\pi(s,a)=\EX_\pi[\sum_{t=0}^{\infty} \gamma^t r_{t+1} | s,a]$. The advantage is $A_\pi(s,a)=Q_\pi(s,a)-V_\pi(s)$. We define the un-discounted occupancy of a state $s$ under policy $\pi$ as,
\begin{equation}
    \mu_{\pi} (s) = \sum_{s_t \in S} \sum_{t=0}^{\infty} \mathbb{P}\big(s_{t+1}=s | s_t, a_t \sim \pi(a_t|s_t)\big).
\end{equation}

Partially observable Markov Decision Processes (POMDPs) are a generalization of Markov Decision Processes (MDPs), where the agent does not have the complete knowledge of the state. It takes an action at a state based on its \emph{observation}, which is an encoding of the underlying state of the environment. One of the earliest works that introduced POMDPs is~\citep{ASTROM1965}, which established the optimal \emph{control} (equivalently the action of the agent) with incomplete information about the state. 

A POMDP can be modeled by a tuple $\langle S, \mathcal{A}, \mathbb{P}, R, \Omega, O  \rangle$, where $\Omega$ is a set of observations that the agent can experience in its \textit{world}. $T: S \times \mathcal{A} \rightarrow \Pi(S)$ is the state-transition function giving for each world state and action, a probability distribution, $\Pi(S)$, over world states.  $O(s',a,o)$ is the probability of observing $o \in \Omega$ if the agent took action $a$ and transitioned to state $s'$.

\section{Related Work}
In this section, we briefly address the works closely related to observational RL.

\textbf{Inverse Reinforcement Learning} (IRL) is an imitation learning method where the agent first recovers the expert's reward function and then learns its own optimal policy using the estimated expert's reward~\citep{ng2000algorithms}. 

In IRL, it is assumed that the expert acts optimally with respect to its reward, i.e., 
\begin{equation}
    \mathbb { E }_{\pi _ { E }} \left[ \sum _ { t } \gamma ^ { t } \hat { r } \left( s _ { t } , a _ { t } \right)\, \right] \geq \mathbb { E }_{\pi} \left[ \sum _ { t } \gamma ^ { t } \hat { r } \left( s _ { t } , a _ { t } \right)\, \right] \quad \forall \pi,
\end{equation}
where $\pi_E$ is the optimal expert policy, $\pi$ is any policy,  $\hat r$ is the expert's reward estimated by the agent.

The agent learns an optimal policy to maximize $\mathbb {E}_{\pi_A} \Big[\sum_{t = 0}^\infty \gamma^t \hat r\ (s_t,a_t) \Big]$,
where $\pi_A$ is the agent's policy. IRL is closely related to observational RL in situations where the agent sees the state-only demonstrations of the expert.

\textbf{Variational Autoencoders} (VAE)~\citep{kingma2013auto} consist of two networks. A \emph{generative network} $p(x|z)$ or $p(s|z)$ for the reconstruction, samples visible reconstructed variables $s$ given latent variables $z$. A \emph{variational inference network} (Encoding network) $q(z|x)$ maps known visible variables $x$ to latent variables $z$.  This approximates a prior distribution $p(z)$.  The objective of VAEs is to maximize the evidence lower bound, $\text{ELBO}(\theta,\phi)=\EX_{q_{\theta}(z|x)}[\log p_{\phi}(x|z)] + D_{\mathrm{KL}}(q_{\theta}(z|x)  \lVert p(z)$).

\textbf{PixelCNN for Count Based Exploration} \citep{NIPS2016_6383}\label{neural_density_prelim} gives the definition of a pseudocount, derived from a density model, to be used in \emph{count-based exploration}. It is computed from a density model $\rho$ over a finite space $\chi$, with $\rho_n(x)$ the probability assigned by $\rho$ to a state $x$ after  training on a sequence of length $n$ of observed states. Using \textit{PixelCNN} \citep{NIPS2016_6527} as a density model \citep{pmlr-v70-ostrovski17a}, a pseudocount is computed and used as an exploration bonus directly on the observed reward in a DQN \citep{Mnih2013PlayingAW}.  It is shown to improve speed of learning in numerous Atari 2600 game environments.

\textbf{Safe Reinforcement Learning} is the problem of learning a policy that maximizes expected return while ensuring that some safety constraints are met. The exploration process in many of these problems is unique in that it incorporates external knowledge of \emph{risky} areas of the state and action space, which can also result in the decrease in training time~\citep{JMLR:v16:garcia15a}. A common algorithm used is constrained policy optimization given a constrained MDP \citep{pmlr-v70-achiam17a} \citep{Altman99constrainedmarkov}.  Other methods have formulated a policy gradient algorithm where the CVaR of the rewards is minimized in a constrained optimization problem \citep{Chow:2014:ACO:2969033.2969218}.

\section{Learning to Avoid Demonstrator State Trajectories}
The agent observes a set of state-trajectories from a bad demonstrator $\mathcal{D}$. Let $\mathcal{T}_\mathcal{D}=\{\tau_1,\tau_2,\dots,\tau_n\}$ be the state trajectories of the demonstrator, $\tau_i \sim \tau_{\mathcal{D}}$ where $\tau_i = \{s_0,s_1,\dots,s_k\}$. Denote $\mathbb{P}_{\tau}$ the distribution of demonstrator trajectories $\tau$. Given the demonstrator trajectories, we estimate the state occupancy distribution of the demonstrator policy by training a VAE on the state occupancy of each demonstrator trajectory (explained below) or we can more simply average the state occupancy measures of all demonstrator trajectories.

To avoid the demonstrator trajectories, we can, as an example, derive an optimization problem based on the discounted sum of rewards. We formulate a scenario where the agent wishes to find a policy that maximizes an arbitrary probability distance metric term $D(\cdot)$ (e.g. KL divergence) capturing the distance between the state occupancy distribution induced by its policy and the one induced by demonstrator policies (the probability of the state being in a demonstrator trajectory).  The dual of the constrained optimization problem is given as follows, with \emph{Lagrange multipliers} $\lambda \geq 0$
\begin{equation}\label{eq:MLE-dual}
\max_\pi \min_{\lambda \ge 0} \EX_{\pi} \Big[\sum_{t=0}^{\infty} \gamma^t r_t \Big] + \lambda D\Big(\EX_{\tau}\Big[\sum_{t=0}^{\infty} \mathbb{P}(s_t=s)\mathbb{P}(s \in \tau)\Big] - \mu_{\pi}(s)\Big).
\end{equation}
We are given demonstrator trajectories and need to compute a stationary distribution, $d_e(s)$, which can be following an optimal policy or a policy we wish to avoid depending on the problem. We can estimate this distribution using a VAE model such that we are building a generative model $p(s|z)$ conditioned on latent variable $Z$. We can then marginalize out latent variables to give $p(s) \approx d_e(s)$: our demonstrator stationary distribution. Similarly, we can train a VAE for the agent learner.  

This method computes a stationary distribution approximation for the agent $d_\pi(s)$ and for the demonstrator $d_e(s)$.  
We can consider metrics for probability distribution such as KL divergence where we would have a general goal of
\begin{equation}
    D_{\mathrm{KL}}(d_{\pi}(s) \lVert d_{e}(s)) = \EX_{s \sim d_{\pi}(s)} \Big[\log {\frac{d_\pi(s)}{d_e(s)}}\Big].
\end{equation}
The agent wishes to consider the KL divergence between the stationary distributions of the demonstrator and the agent in the case of avoidance.  A policy optimization objective function, over the discounted sum of rewards, to maximize is therefore
\begin{equation}
    J(\pi) = \EX_{\pi} \Big[\sum_{t=0}^{\infty} \gamma^t r_t + \beta D_{\mathrm{KL}}(\hat d_{\pi}(s) \lVert \hat d_{e}(s))\Big],
\end{equation}
where $\hat d_{\pi}$ and $\hat d_e$ are the estimates of the state occupancy measures of the agent (under policy $\pi$) and the expert (demonstrator) respectively. We compute them by taking a finite number of uniformly distributed samples from the two VAEs (as a generative model) and then finding the mean state occupancy measures for each. 
If we have multiple demonstrators, we can compute for each one a distance metric term with respect to the current policy. We also select a parameter $\beta$ as a coefficient for the distance metric in $J$. 
An example of an avoidance learning algorithm which maximizes KL divergence between state occupancy distributions with Proximal Policy Optimization (PPO) \citep{Schulman2017ProximalPO} and an arbitrary advantage estimation method (ie. Generalized Advantage Estimation (GAE)) \citep{Schulmanetal_ICLR2016} is shown in Algorithm 1. If we wish to perform this algorithm without using a VAE, we simply average the state occupancy measures from the demonstrator trajectories and the corresponding state occupancy measures calculated from trajectories sampled from the policy in Line 6.
\renewcommand{\algorithmicrequire}{\textbf{Input:}}
\begin{algorithm}
\caption{PPO with avoidance KL term (using VAE)}
 \begin{algorithmic}[1]
\REQUIRE $\text{Demonstrator Trajectories: }\mathcal{T}_\mathcal{D}, \text{Initial Policy Parameters: }\theta_0, \text{KL Coefficient: }\beta$
\STATE $ \mu_{e} (s) \leftarrow \EX_{\tau \sim \mathcal{T}_\mathcal{D}}[\sum_{s_t \in \tau} \sum_{t'=0}^{\infty} \mathbb{P}[s_{t+1}=s | s_t]]$
\STATE $\text{VAE}_{e} \leftarrow trainVAE(\mu_e (s))$
\STATE $d_{e}(s) \sim sample(\text{VAE}_{e})$
\FOR{$k=0,1,2,\dots$}
\STATE $\pi_{k} \leftarrow \pi(\theta_k)$
\STATE $\mu_{\pi_k} (s) \leftarrow \sum_{s_t \in S} \sum_{t'=0}^{\infty} \mathbb{P}[s_{t+1}=s | s_t, \pi_{k}(s_t)]$
\STATE $\text{VAE}_{\pi_k} \leftarrow trainVAE(\mu_{\pi_k} (s))$
\STATE $d_{\pi_k}(s) \sim sample(\text{VAE}_{\pi_k})$
\STATE $\text{Collect advantage estimates } \hat{A}^{\pi_k}_{t}\text{ using any advantage estimation method}$
\STATE $\hat{A}^{\pi_k}_{t} \mathrel{+}= \beta D_{\mathrm{KL}}(d_{\pi_k}(s) \lVert d_{e}(s))$
\STATE $\theta_{k+1} \leftarrow \operatorname*{arg\,max}_\theta \EX_{\tau \sim \pi_k}[\sum_{t=0}^{\infty} \text{min}(r_t(\theta)\hat{A}^{\pi_k}_{t},\text{clip}(r_t(\theta),1-\epsilon,1+\epsilon)\hat{A}^{\pi_k}_{t})]$
\ENDFOR
\end{algorithmic}
\end{algorithm}
\section{Using Neural Density Models For an Avoidance Bonus}\label{neural_density}

In this section, our motivation is to propose a solution for AvL in partially observable environments where the agent receives raw images as observation. The previous approach of estimating state occupancy distribution is not tractable anymore: since the environment is partially observable, two scenarios are possible depending on if we have access or not to the agent's coordinates. In the case we have them, the agent can experience multiple points of view for the same 3D Cartesian coordinates, depending on the angle of observation. The number of possible states would therefore grow. If we can't access the exact positions like in many real-world scenarios, our previous approach of using KL divergence for estimating two states distributions cannot stand anymore.  

Similar to \citep{pmlr-v70-ostrovski17a}, we propose the solution to train a density model $\rho_n(x)$ as described in Section \ref{neural_density_prelim} on the sequence of observations ($x_1,\dots, x_n \sim \tau_{i}$) with $\tau_i \in \mathcal{T}_\mathcal{D}$ for a POMDP. 

Let $\mathcal{X}$ be the space of observable states induced by any agent and its training environment.  Empirical distribution $F_{\mathcal{T}_\mathcal{D}}(\mathcal{X})$ is induced by the demonstrator $\mathcal{D}$'s observations. For $x,y \in \mathcal{X}$ with $F_{\mathcal{T}_\mathcal{D}}(x) > F_{\mathcal{T}_\mathcal{D}}(y)$, their corresponding neural density model estimates would show $\rho_n(x) \gtrapprox \rho_n(y)$. We wish to avoid observing states $x$ more than $y$ in our agent training to avoid the demonstrator trajectories. 

We now define the pseudocount and exploration bonus. $\rho'_{n}(x)$ is the probability that $\rho_n$ would assign to $x$ if it was trained on $x$ again. Given the density model $\rho$, we can compute a prediction gain of the model $\text{PG}_n(x) = \text{log} \rho'_{n}(x) - \text{log} \rho_{n}(x)$.  The prediction gain is also enforced by a threshold to be learning positive $(\text{PG}(x))_{+}$. $\rho_n(x)$ is learning positive if $\rho'_{n}(x) \geq \rho_{n}(x)$ for all $x_1,\dots x_n,x \in \chi$.

We use a Gated PixelCNN based density model to generate a pseudocount defined by:
\begin{equation}
\hat{N}_n(x) = \left(\text{exp} \left(c \cdot n^{-1/2} \cdot (\text{PG}_n(x))_{+}\right)-1 \right)^{-1}
\end{equation}
with $c \cdot n^{-1/2}$ corresponding to a prediction gain decay. We obtain $\hat{N}_{n,\mathcal{D}}(x)$ by training the density model on trajectories sampled from the demonstrator. We can say this is a density estimate of the demonstrator observations in a POMDP. We then have an \textit{avoidance} reward bonus that we define similarly to an \textit{exploration bonus} at step $n$,
\begin{equation}
    r_{+}(x) = \beta (\hat{N}_{n,\mathcal{D}}(x))^{-1/2},
\end{equation}
which is then added to observed agent rewards to encourage avoidance of the states with high frequency in demonstrator trajectories. This bonus is low for demonstrator observations and higher for observations not commonly occupied by the demonstrator. We therefore use the pseudocount as a reward bonus for the agent during training, since it will give positive feedback for exploring the states spaces not seen by the demonstrator and thus avoid it. 
with $\beta$ being the \textit{pseudocount weight} coefficient parameter. We expect this bonus to be useful during training for environments with sparse rewards, because it now provides feedback at every step to the agent.
\section{Experiments}
We investigate the following questions with our experiments:
\begin{itemize}
\item Does using a KL penalty or pseudocount avoidance bonus actually enforce safer trajectories during training? Does it compare with baseline Safe RL methods?
\item Is sample efficiency improving in environments with sparse rewards while avoiding unsafe regions?
\end{itemize}
We use the below task settings to explore these questions. Our experiments are designed to emulate realistic control situations; such as avoiding dangerous regions in 2D fully observable grid-world environments and 3D partially observable worlds, with sparse rewards, selecting the right path/room to reach the goal, and selecting the right objects. All are implemented with Gym supported environments and the hyperparameters used in our experiments are described in Appendix C.2.  

\subsection{2D Grid-world Environments}
We build our 2D grid-world environments using the Gym "MiniGrid" package \citep{gym_minigrid}. The environments are fully observable and each observation is an (w, h, 3) tensor. At each timestep, the agent can change its direction, actions are as follows: \textit{turn-left}, \textit{turn-right}, \textit{move-forward}. Facing a wall, the agent will stay in the same state if it moves forward into the wall. 
Rewards are sparse: we gave a non-zero reward to the agent only when it fully completed the mission, and the magnitude of the reward was $1 - 0.9 \cdot n / n_{\text{max}}$, where $n$ is the length of the successful episode and $n_\text{max}$ is the maximum number of steps that we allowed for completing the episode, different for each mission. If the agent goes into lava or reaches the maximum number of steps authorized for each episode, the episode ends with 0 reward. 
Some examples are shown in Figure \ref{fig:grid_envs} (see Appendix A for more details about MiniGrid).

We train the demonstrator to go to lava which is the dangerous behavior for the agent. In case of multiple cells of lava to avoid, we can train several demonstrators, each one being trained to go to a different lava cell. 

For experiments, we used the PPO algorithm with parallelized data collection and GAE. Each environment is run with 20 random network initializations.

When executing Algorithm 1, PPO with avoidance KL penalty, we sample 100 trajectories from the demonstrators' policies.  We then use these trajectories to estimate the demonstrator state occupancy distribution with 10,000 samples from the VAE trained on the demonstrator state occupancies.


\vspace*{-3mm}
\begin{figure}[H]%
    \centering
        \subfloat[SuperGrid ]{{\includegraphics[width=3cm]{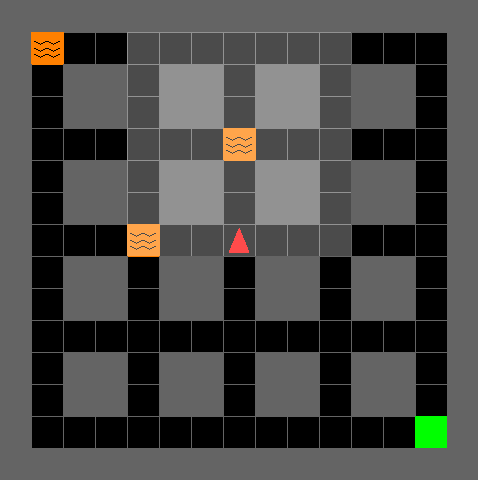}}}%
        \vspace{0.3cm}
            \subfloat[LavaAvoidance ]{{\includegraphics[height=3cm]{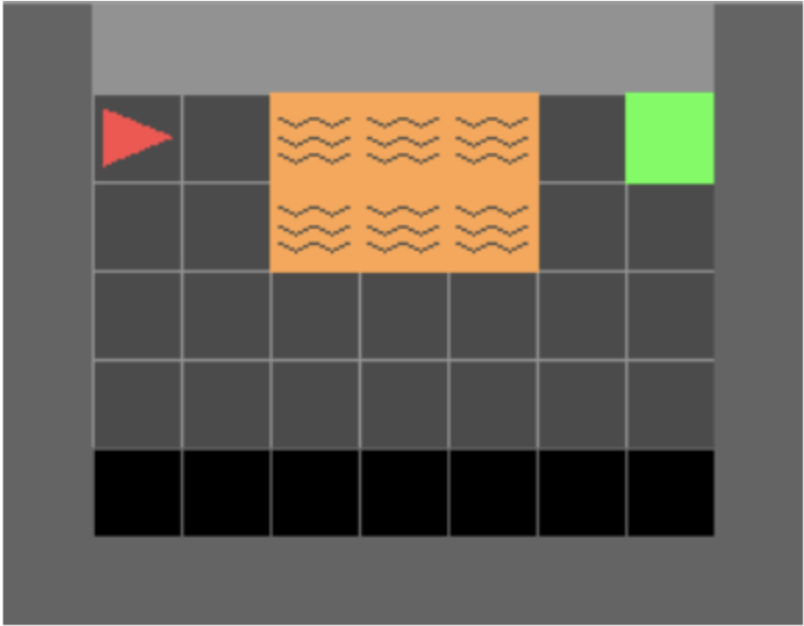}}}%
                    \vspace{0.3cm}
    \subfloat[ThinHallwaysGrid ]{{\includegraphics[height=3cm]{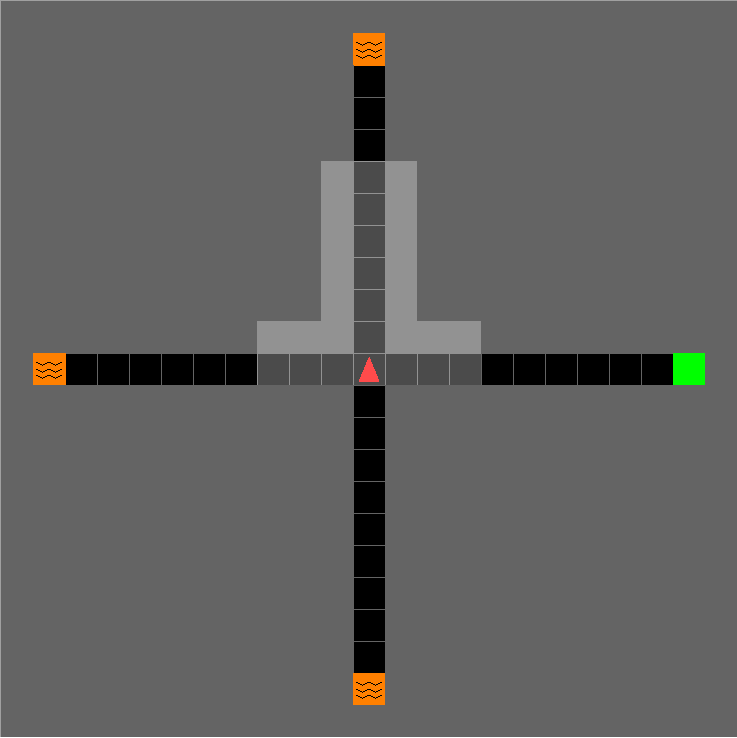}}}%
                \vspace{0.3cm}
    \subfloat[SuperFlowerGrid ]{{\includegraphics[height=3cm]{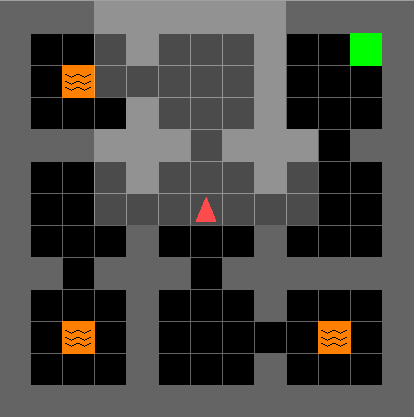}}}
    \vspace*{-10mm}
    \caption{2D Grid Environments. The Demonstrator(s) policy is trained with a goal at each lava cell.}%
    \label{fig:grid_envs}%
\end{figure}
\vspace*{-3mm}
\subsection{3D Partially observable environments}

We use the Gym "MiniWorld" package \citep{gym_miniworld} to create 3D environments with an egocentric point of view for the agent. Examples of these environments are presented in Figure \ref{fig:world_envs} (see Appendix B for more information on MiniWorld). We can solve these environments using PPO with a convolutional actor-critic architecture (see Appendix C.1 for more details).

As explained previously with MiniGrid, we train the demonstrators modifying the reward function so that the demonstrator will learn the behavior we later want to avoid for safety reasons. For example, on \textit{Sidewalk} the demonstrator will receive a reward if it goes to the street or walks along with it, or even if it gets stuck in front of a wall.

On the other hand, when training our true agent to go to the goal (the red box), we don't want it to go to the street where there is a potential danger. Therefore, we aim to give this agent an avoidance bonus at each step, using the neural density model described in Section \ref{neural_density}. The agent receives a bonus when seeing observations far from the demonstrator distribution. This incentivizes the agent to remain far from danger, and would additionally improve sample efficiency because the rewards would not be sparse anymore.

\begin{figure}[H]%
    \centering
    \subfloat[Sidewalk: the demonstrator walks along the street or goes in it.  ]{{\includegraphics[width=3.5cm]{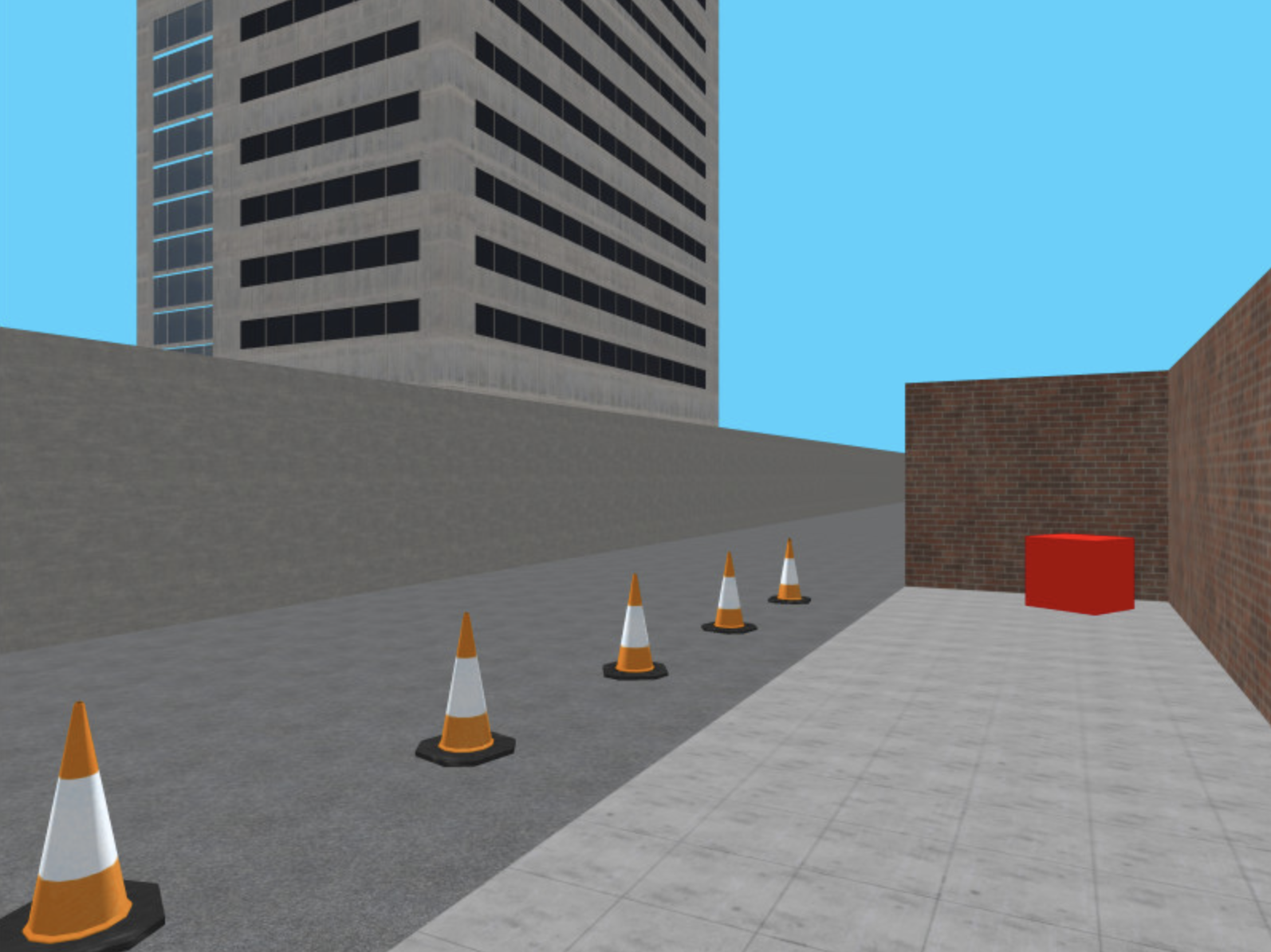}}}%
                \vspace{0.5cm}
    \subfloat[FourRooms: the demonstrator goes to the yellow box instead of the red box. ]{{\includegraphics[width=3.5cm]{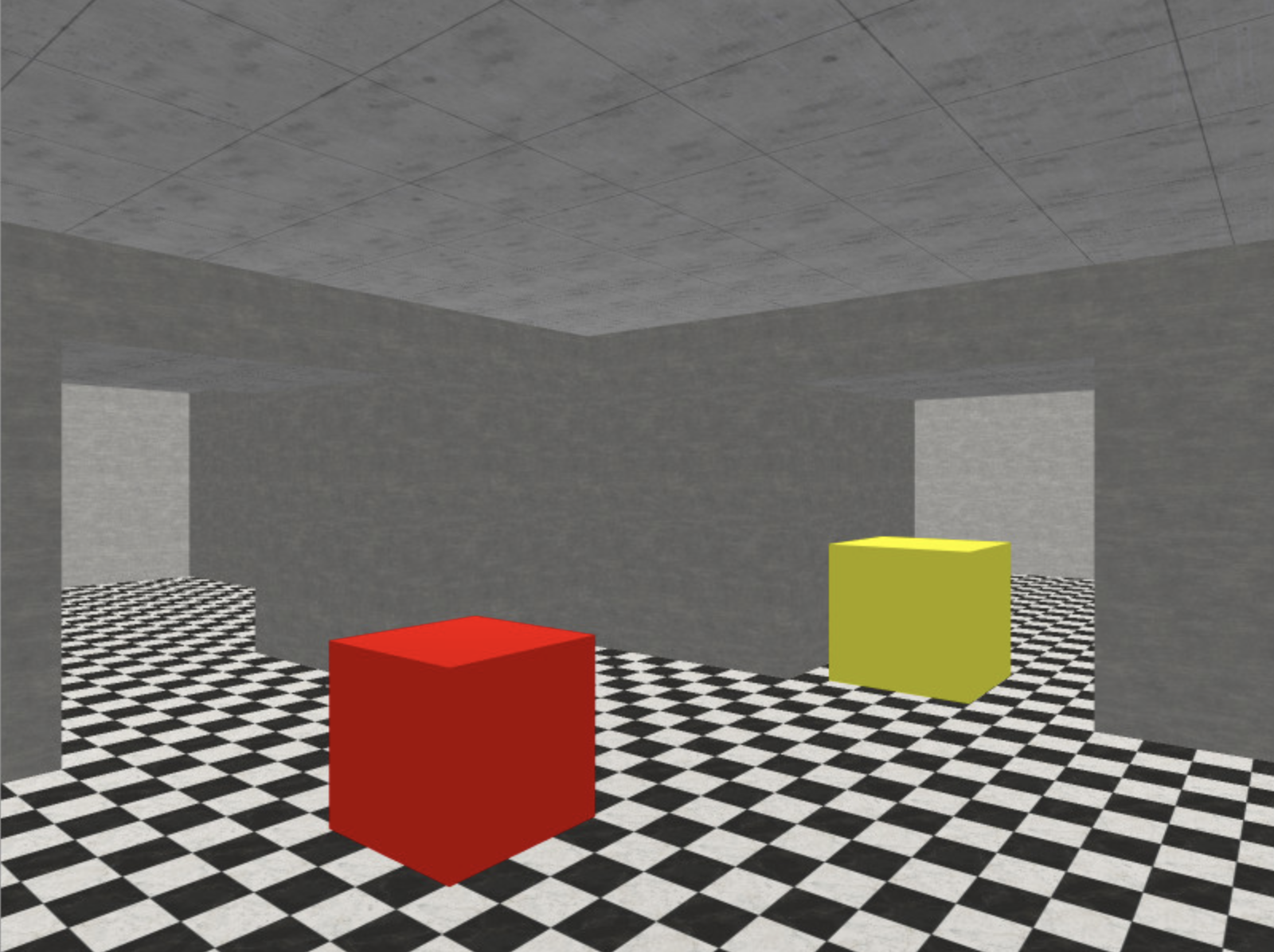}}}%
                \vspace{0.5cm}
        \subfloat[TMaze (Overhead view): the demonstrator goes to the hall without the goal. ]{{\includegraphics[width=3.5cm]{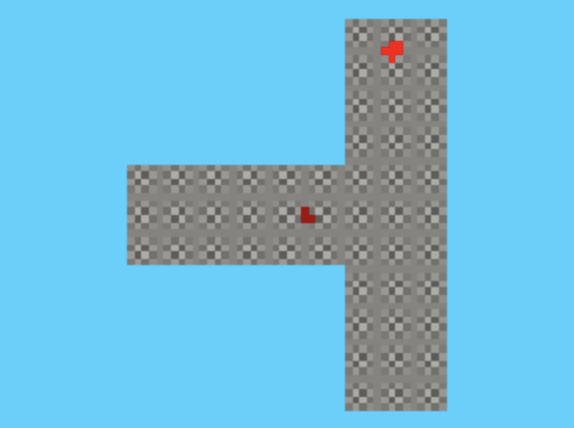}}}%
    \caption{3D Partially Observable Environments}%
    \label{fig:world_envs}%
\end{figure}
\subsection{Results for 2D Grid-world environments}
\begin{figure}[H]\label{fig:example}%
    \centering
    \subfloat[ThinHallwaysGrid without the AvL term ]{{\includegraphics[width=2.5cm, height=2.5cm]{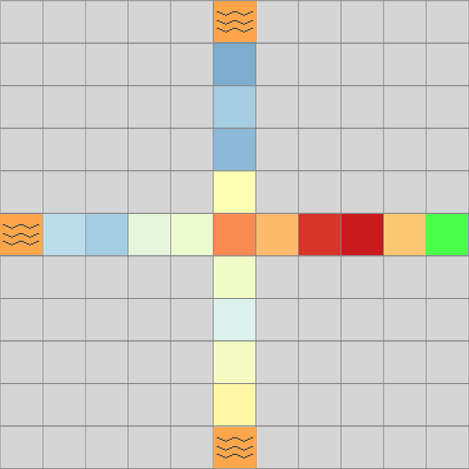}}}\hfil%
    \subfloat[ThinHallwaysGrid with AvL term]{{\includegraphics[width=2.5cm, height=2.5cm]{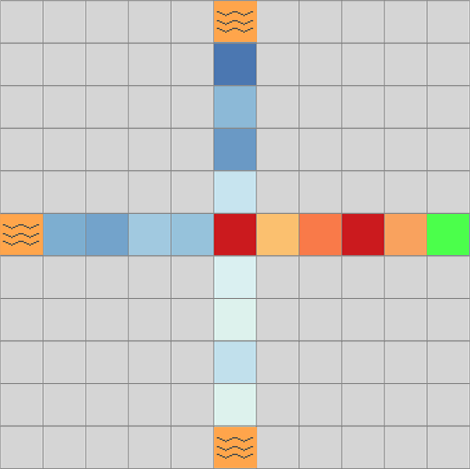}}}\hfil%
    \subfloat[LavaAvoidance without the AvL term ]{{\includegraphics[ width=2.5cm, height=1.8cm]{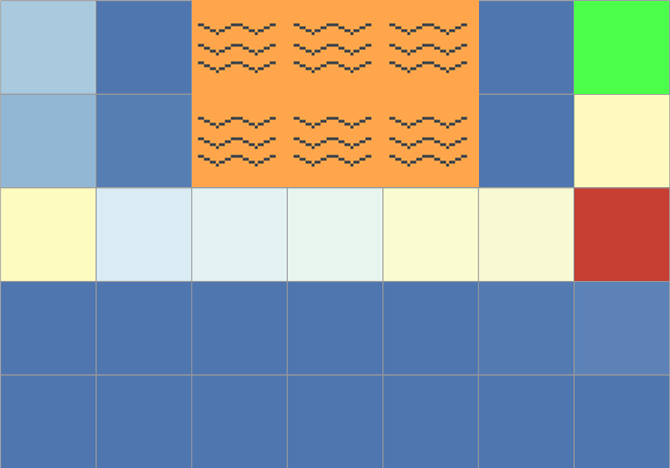}}}\hfil%
    \subfloat[LavaAvoidance with AvL term  ]{{\includegraphics[ width=2.5cm, height=1.8cm]{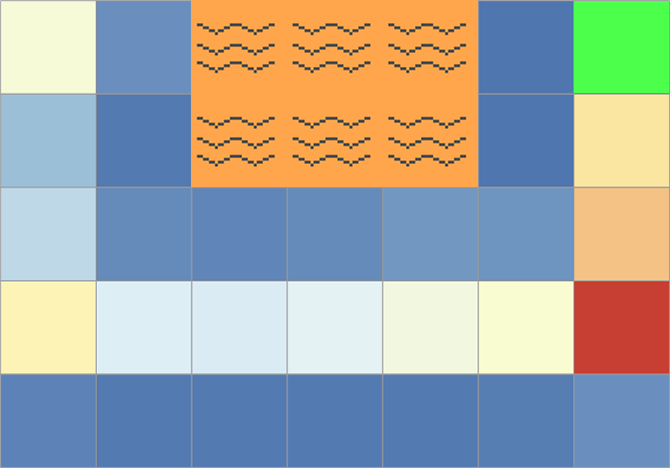}}}\hfil%
        \subfloat[LavaAvoidance with PPO-CVaR  ]{{\includegraphics[ width=2.5cm, height=1.83cm]{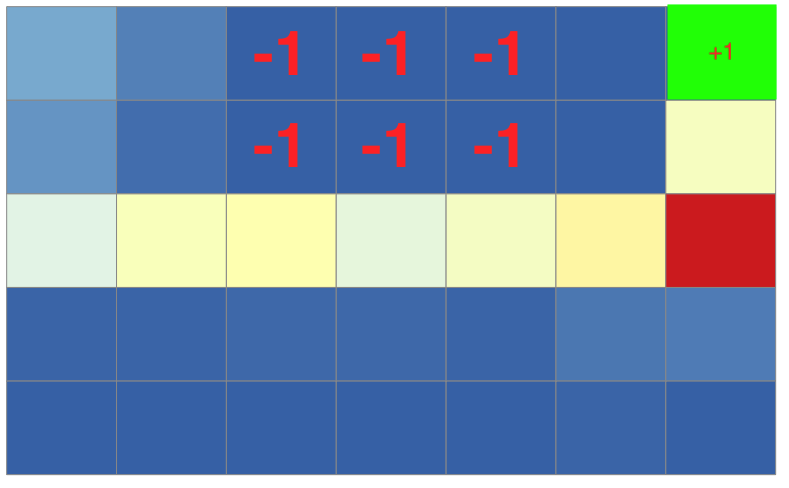}}}\hfil%
    \caption{The state occupancy distributions induced by each model's policy. Distributions are obtained after 10 updates for ThinHallwaysGrid, at convergence for AvoidLava. Red is higher state occupancy probability.}%
        \label{fig:ssreps}%
\end{figure}

The avoidance learning method described in Section 4 is shown to have decent level of increases on the sample efficiency of PPO but also generates policies that show more significant avoidance of dangerous regions. In relatively simple Grid environments, we did not expect large increases in sample efficiency as the solutions are relatively easy in a small state-space compared to complex 3D environments. In Figure \ref{fig:ssreps} the state occupancy distribution show that the introduction of the KL term during training leads to convergence of a policy that is one cell further away from the lava cells.  (a,b) show faster convergence in terms of sample efficiency of the policy towards the goal and avoidance of dangerous states during training.

In addition, we compared against a trajectory-based policy gradient CVaR optimization method that is a variation of PPO (PPO-CVaR) (Described in Appendix D.1) and the original version presented in \citep{Chow:2014:ACO:2969033.2969218} (PG-CVaR) (which uses the discounted sum of rewards as the policy gradient score function).  For the CVAR experiments, we introduced a reward of $-1$ to each lava cell instead of ending the episode.  
    \vspace{-5mm}
    \vspace{-5mm}
In Table \ref{fig:sampling}, we see that introducing the KL term leads to faster policy convergence (for the optimal hyperparameters) in all Grid environments tested.  We compare using the VAE to estimate the stationary distributions of the state occupancy vs. averaging the state occupancies computed from sampled trajectories/demonstrator trajectories. The value of the KL term depends on the extent of divergence from the demonstrator trajectories and therefore has a large effect on convergence.  For some environments, the agent temporarily follows the trajectory of the demonstrator and eventually diverges from it to achieve the goal. By varying the KL weight, we find a suitable KL term to achieve \textit{sufficient} divergence while still accomplishing the task.  The method performs best in environments where there is a separated area (e.g. a room with one entrance) where the lava is present in.  The fact that using AvL without training a VAE either performs worse or only marginally better than simple averaging to estimate the state occupancy distributions indicates this computationally expensive step could be unnecessary given sufficient demonstrator trajectories. 

\begin{table}[H]
  \caption{Average number over 20 seeds of samples needed to reach 85\% success for MiniGrid environments (numbers in thousand of frames). AvL (VAE) indicates that we used PPO with avoidance learning (KL term) and estimated the state occupancy distributions using a VAE, AvL (Avg) indicates that we simply averaged the sampled trajectory state occupancy measures.}
  \vspace{5mm}
  \label{sample-table}
  \centering
  \begin{tabular}{llllll}
    \toprule
    \cmidrule(r){1-2}
    Environment     & AvL (VAE) & AvL (Avg) & PPO & PPO-CVaR & PG-CVaR    \\
    \midrule
    ThinHallwaysGrid & \textbf{25.82} & 37.13   & 27.34 & 25.99 &  2439.42   \\
    SuperFlowerGrid    & 640.41 & \textbf{304.24}  & 972.33 & 855.42 & 4091.95    \\
    SuperGrid    & 162.22 & \textbf{106.49}    & 177.97 & 169.13 & 3822.70  \\
    LavaCrossingGrid-1    & \textbf{842.34} & 854.75    & 890.54 & 893.78 & 3830.41  \\
    LavaCrossingGrid-2    & \textbf{418.13} & 422.50   & 436.33 & 451.91 & 3545.78 \\
    LavaAvoidanceGrid    & 67.58 & \textbf{48.02}   & 69.63 & 65.37 & 1822.35  \\

    \bottomrule
  \end{tabular}
  \label{fig:sampling}
\end{table}

\subsection{Results for 3D environments}

\begin{figure}[!h]%
    \centering
    \includegraphics[width=14cm]{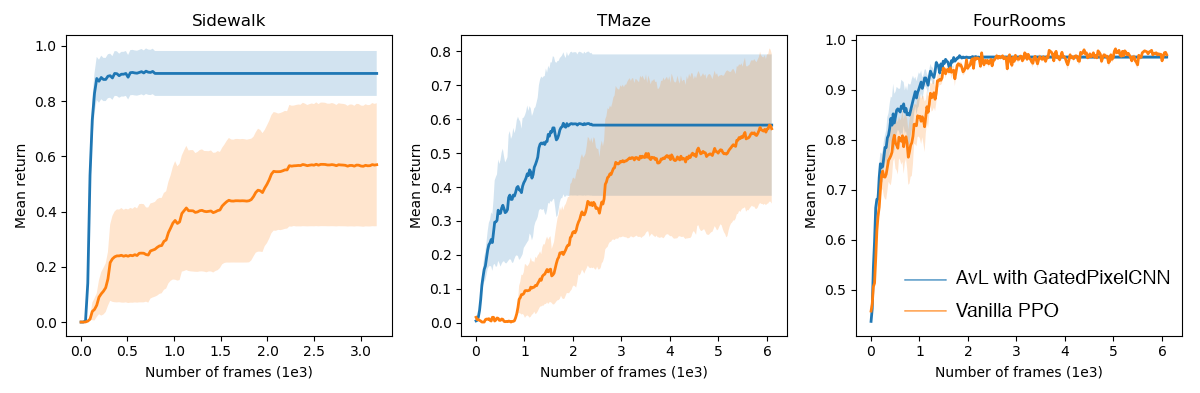}
    \caption{Success rate w.r.t to the number of frames for MiniWorld environments. }%
    \label{fig:samplingWorld}%
\end{figure}

For 3D MiniWorld environments we see remarkable improvements in both achieved success rate and number of frames to reach convergence. Variance is also reduced. (Figure \ref{fig:samplingWorld}).

To explain this incredible improvement, we call for caution and wish to point out the importance of the choice of the seeds to run the experiments. We randomly selected 20 to run our experiments. In addition, the road is a very large region (the entire left side of the world) and entering it ends the episode.  By learning to avoid this region, the agent is confined to a relatively small sidewalk and can reach the goal more easily. 

Hyperparameters are the same for experiments using PPO with and without pseudocount bonus and we optimized them to have the best training curve, in terms of number of frames until convergence, possible for experiments using PPO without bonus. We could witness that for some seeds, the policy did not converge towards succeeding in the missions, while it did with the pseudocount bonus. We hypertuned the pseudocount weight parameter, as we previously did with the KL weight parameter and achieved the best results with 0.1. 

Regarding safety during training, the agent indeed avoids going on the road on the Sidewalk environment and travels less frequently to the incorrect box on FourRooms.
\section{Discussion}
We propose a novel algorithm for Avoidance Learning through observational reinforcement learning as well as a novel method for observational reinforcement learning in partially observable continuous environments. We demonstrate that learning from observation has the ability to learn safer policies, provide a safer learning process and learn these policies more efficiently.  The approach does not require any explicit knowledge of demonstrator actions, any engineering of negative rewards, or known policy constraints. State-only demonstrations are sufficient. The results of this method are of interest to autonomous vehicles, as they can be trained on cars without an action recording apparatus.  Simple observations of poor driving behavior can be used via simple video recordings.  

Some limitations can arise from our approach. A fundamental assumption for this to work is that one can sample state trajectories from the demonstrator. In addition, estimating stationary state distributions or using GatedPixelCNN is also computationally expensive and in the case of the VAE, it requires training at every update: we trade computation time for sample efficiency and safety. 
\section{Acknowledgements}
We thank Maxime Chevalier-Boisvert, Riashat Islam, David Yu-Tung Hui, Dzmitry Bahdanau and Charles Guille-Escuret for helpful discussions. We thank Jordan Hoffman, Vincent Luczkow, Guillaume Alain for their help in reviewing the paper.

\newpage
\bibliographystyle{plainnat}

\bibliography{biblio}

\newpage
\appendix

\section{MiniGrid Environments for OpenAI Gym\label{sec:minigrid}}

MiniGrid, is an open source gridworld package\footnote{https://github.com/maximecb/gym-minigrid} that includes a family of reinforcement learning environments compatible with the OpenAI Gym framework. Many of these environments are customizable so that the task difficulty can be adjusted (e.g., the size of rooms, the number and type of objects, the topology).

\subsection{The World}

In MiniGrid, the world is a grid of size $w \times h$. Each tile in the grid contains exactly zero or one object. We use the objects: wall, lava and goal. Each object has an associated discrete color, which can be one of red, green, blue, purple, yellow and grey. By default, walls are always grey and goal squares are always green.

\subsection{Reward Function}

Rewards are sparse for all MiniGrid environments. Episodes are terminated with a positive reward when the agent reaches the specified goal (generally the green goal square). Otherwise, episodes are terminated with zero reward when a time step limit is reached or the agent goes into lava.

The formula for calculating positive sparse rewards is $1 - 0.9 * (step\_count / max\_steps)$. That is, rewards are always between zero and one; the quicker the agent successfully completes an episode, the closer the reward is to $1$. The $max\_steps$ parameter is different for each environment, and varies depending on the size of each environment, with larger environments having a higher time step limit.

\subsection{Action Space}

There are seven actions in MiniGrid:  turn left, turn right, move forward, pick up an object, drop an object, toggle and done. For the purpose of this paper, the pick up, drop, toggle and done actions are irrelevant. The agent can use the \textit{turn-left} and \textit{turn-right} actions to rotate and face one of 4 possible directions (north, south, east, west). The \textit{move forward} action makes the agent move from its current tile onto
the tile in the direction it is currently facing, provided there is nothing on that tile, or that the tile contains an open door.
The agent can open doors if they are right in front of it by using the toggle action.

\subsection{Observation Space}

We are using fully observable grids of size $w\times h$. The observations are provided as a tensor of shape (w, h, 3). However, note that these are not RGB images. Each tile is encoded using 3 integer values: one describing the type of object
contained in the cell, one describing its color, and a flag indicating whether doors are open or closed. This compact encoding was chosen for space efficiency and to enable faster training. The fully observable RGB image view of the environments shown in this paper is provided for visualization.

\section{MiniWorld Environments for OpenAI Gym}\label{miniworld}

MiniWorld \footnote{https://github.com/maximecb/gym-miniworld} \citep{gym_miniworld} is a minimalistic 3D interior environment simulator for reinforcement learning and robotics research. It can be used to simulate environments with rooms, doors, hallways and various objects (e.g: office and home environments, mazes).

\subsection{The World}

In MiniWorld, the world is made of static elements (rooms and hallways), as well as objects which may be dynamic, which we call entities. Environments are essentially 2D floorplans made of connected rooms. Rooms can have any convex outline defined by at least 3 points. Portals (openings) can be created in walls to create doors or windows into other rooms.

\subsection{Coordinate System}

MiniWorld uses OpenGL's right-handed coordinate system. The ground plane lies along the X and Z axes, and the Y axis points up. When direction angles are specified, a positive angle corresponds to a counter-clockwise (leftward) rotation. Angles are in degrees for ease of hand-editing. By convention, angle zero points towards the positive X axis.

\subsection{Observations}

The observations are single camera images, as numpy arrays of size (80, 60, 3). These arrays contain unsigned 8-bit integer values in the [0, 255] range.

\subsection{Actions}

For simplicity, actions are discrete. The actions we use are: \textit{turn-left}, \textit{turn-right} and \textit{move-forward}.
The turn and move actions will rotate or move the agent by a small fixed interval. The simulator assumes that the agent is a differential drive robot.

\subsection{Reward function}

Each environment has an associated max-episode-steps variable which specifies the maximum number of time steps allowed to complete an episode. By default, rewards are sparse and in the [0, 1] range, with a small penalty being given based on the number of time steps needed to complete the task: $1 - 0.9 * (step\_count / max\_steps)$. If the task is not completed within the maximum number authorized, a reward of 0 is produced.

\section{Hyperparameters and Models Architectures}

\subsection{Architecture of the policies}

We rely on the actor-critic architecture on both MiniGrid and MiniWorld.

\begin{figure}[H]%
    \centering
    {{\includegraphics[width=12.5cm]{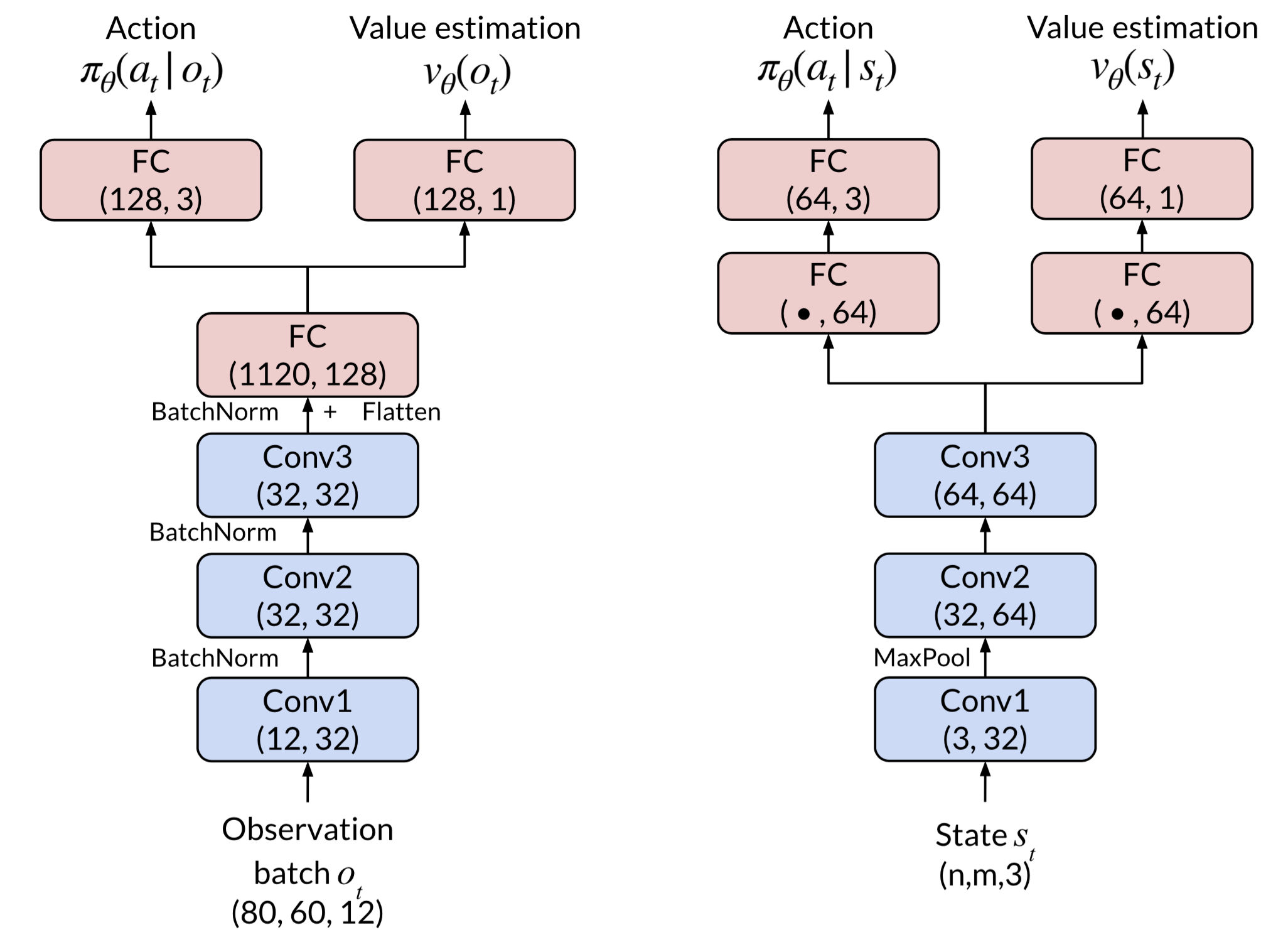}}}\hfil%
    \caption{Architectures of the actor-critic policies on MiniWorld (left) and MiniGrid (right). }%
    \label{fig:architectures}%
\end{figure}
\subsection{Hyperparameters used in MiniGrid and MiniWorld}\label{hyper}

\begin{table}[tbh]
    \caption{Shared parameters for benchmark tasks in MiniGrid}
  \begin{center}
    \begin{tabular}{lr}
      \toprule
      Parameter & Value  \\
      \midrule
      entropy coefficient
      & $10^{-2}$\\
      value loss coefficient & 0.5\\
            discount                                        & 0.99      \\

      maximum norm of gradient in PPO &0.5\\
      number of PPO epochs &4\\
      batch size for PPO &256\\
      entropy coefficient                    & $10^{-2}$      \\
      clip parameter &0.2
             \\
    
      \bottomrule
    \end{tabular}
  \end{center}
\end{table}

\begin{table}[tbh]
    \caption{Shared parameters for benchmark tasks in MiniWorld}
  \begin{center}
    \begin{tabular}{lr}
      \toprule
      Parameter & Value  \\
      \midrule
      value loss coefficient in PPO & 0.5\\
            discount                                        & 0.99      \\

      maximum norm of gradient in PPO &0.5\\
      number of PPO epochs &4\\
      batch size for PPO &128\\
      entropy coefficient                    & $10^{-2}$      \\
      clip parameter in PPO &0.2
             \\
    $c$ parameter in GatedPixelCNN & 0.1\\
    
      \bottomrule
    \end{tabular}
  \end{center}
\end{table}

\section{Methodology}

\subsection{CVaR Constrained Policy Gradient Method}
(Changed method for new results)
We propose the CVaR constraint for the PPO algorithm. $\mathcal{L}_{\theta_t}^{\text{CLIP}}(\theta)$ is the standard PPO clipped loss. $H_{\alpha}(Z,\upsilon)= \upsilon + \frac{1}{1-\alpha}\EX[(Z-\upsilon)^+]$, with $Z$ being an arbitrary probability distribution. $D_{\pi_{\theta}}(s_0)$ refers to the expected distribution of the discounted sum of rewards starting for state $s_0$ under policy $\pi_\theta$ and is estimated as $\EX_{\tau \sim \pi_\theta}[\sum_{t=0}^{\infty}\gamma^t r_t]$.
\begin{equation}
    \mathcal{L}^{\text{CVaR-PPO}} (\theta) = \mathcal{L}_{\theta}^{\text{CLIP}}(\theta) - \lambda(H_{\alpha}(D_{\pi_{\theta}}(s_0),\upsilon) - \beta)
\end{equation}

\section{Additional environments and results}

\begin{figure}[H]%
    \centering
    \subfloat[LavaCrossingGrid-1 ]{{\includegraphics[width=3cm]{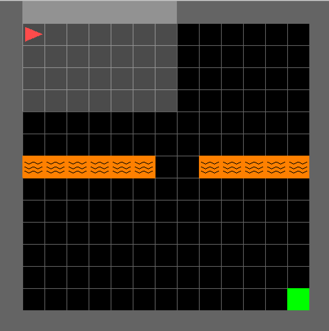}}}%
                        \vspace{0.3cm}
    \caption{Additional 2D Grid Environments. }%
    \label{fig:grid_envs_add}%
\end{figure}

We also show the effect of selecting varying coefficients for the neural avoidance bonus.

\begin{figure}[H]%
    \centering
    \includegraphics[width=5cm]{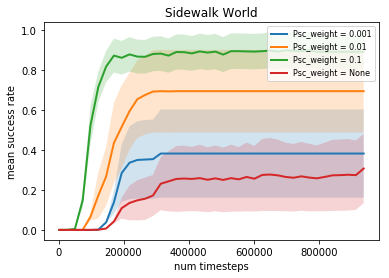}\hfil%
    \caption{Plot of the Success rate of the Sidewalk env with different Pseudo count (PSC) Coefficient values.}%
    \label{fig:samplingWorld_add}%
\end{figure}


\end{document}